\documentclass[conference]{IEEEtran}
\IEEEoverridecommandlockouts
\usepackage{cite}
\usepackage{amsmath,amssymb,amsfonts}
\usepackage{algorithmic}
\usepackage{graphicx}
\usepackage{textcomp}
\usepackage{xcolor}
\usepackage{booktabs}
\usepackage{tcolorbox}
\usepackage{hyperref}
\usepackage{lipsum} 
\usepackage{array}
\usepackage{tabularx}
\usepackage{url}
\usepackage{booktabs} 
\usepackage{multirow}
\usepackage{fontawesome5}
\usepackage[table]{xcolor} 
\usepackage{colortbl}
\usepackage{multirow}
\usepackage{adjustbox}
\usepackage{placeins}

\definecolor{correctgreen}{RGB}{0, 150, 0} 
\definecolor{wrongred}{RGB}{200, 0, 0}
\definecolor{headergray}{gray}{0.9}

\def\BibTeX{{\rm B\kern-.05em{\sc i\kern-.025em b}\kern-.08em
    T\kern-.1667em\lower.7ex\hbox{E}\kern-.125emX}}
\begin{document}

\title{Brain3D: Brain Report Automation via Inflated Vision Transformers in 3D}

\author{\IEEEauthorblockN{Mariano Barone$^a$, Francesco Di Serio$^a$, Giuseppe Riccio$^a$, Antonio Romano$^a$, \\ Marco Postiglione$^b$, Antonino Ferraro$^c$~\IEEEmembership{Member,~IEEE}, Vincenzo Moscato$^a$}
\IEEEauthorblockA{\textit{University of Naples Federico II, Department of Electrical Engineering and Information Technology}, Italy \\
\textit{Northwestern University, Dept. of Computer Science, McCormick School of Engineering and Applied Science}, United States \\ 
\textit{Pegaso University, Department of Information Science and Technology}, Italy \\ 
$^a$\{mariano.barone, francesco.diserio, giuseppe.riccio3, antonio.romano5, vincenzo.moscato\}@unina.it \\
$^b$\{marco.postiglione\}@northwestern.edu, $^c$\{antonino.ferraro\}@unipegaso.it}
}

\maketitle

\begin{abstract}
Current medical vision-language models (VLMs) process volumetric brain MRI using 2D slice-based approximations, fragmenting the spatial context required for accurate neuroradiological interpretation. We developed \textbf{Brain3D}, a staged vision-language framework for automated radiology report generation from 3D brain tumor MRI. Our approach inflates a pretrained 2D medical encoder into a native 3D architecture and progressively aligns it with a causal language model through three stages: contrastive grounding, supervised projector warmup, and LoRA-based linguistic specialization. Unlike generalist 3D medical VLMs, \textbf{Brain3D} is tailored to neuroradiology, where hemispheric laterality, tumor infiltration patterns, and anatomical localization are critical. Evaluated on 468 subjects (BraTS pathological cases plus healthy controls), our model achieves a Clinical Pathology F1 of 0.951 versus 0.413 for a strong 2D baseline while maintaining perfect specificity on healthy scans. The staged alignment proves essential: contrastive grounding establishes visual-textual correspondence, projector warmup stabilizes conditioning, and LoRA adaptation shifts output from verbose captions to structured clinical reports\footnote{Our code is publicly available for transparency and reproducibility: \url{https://github.com/PRAISELab-PicusLab/BrainGemma3D}}.
\end{abstract}


\section{Introduction}
\label{sec:introduction}

Automated radiology report generation has advanced rapidly with the emergence of large vision-language models (VLMs). Systems such as Med-Flamingo \cite{moor2023med}, LLaVA-Med \cite{liu2023visual} and MedGemma \cite{sellergren2025medgemma} demonstrate strong descriptive capabilities; however, their clinical reliability remains limited by hallucinations and weak adherence to diagnostic structure.

In neuro-oncology, this limitation is amplified by a fundamental \emph{volumetric gap}. Brain MRI interpretation, particularly FLAIR, requires coherent 3D spatial reasoning to assess tumor infiltration, hemispheric laterality, and periventricular signal changes \cite{Filippi2016MAGNIMS}. Yet most medical VLMs operate natively on 2D images, resorting to slice-wise decomposition when processing volumetric scans. This strategy disrupts spatial continuity and frequently leads to lateralization errors and false lesion attribution \cite{Bink2018StructuredReporting}. Recent 3D multimodal models (e.g., Med3DVLM \cite{med3dvlm}, M3D-LaMed \cite{m3d_lamed}) introduce native volumetric encoders. However, these systems are typically trained as generalist assistants across heterogeneous modalities and lack domain-specific grounding for neuroradiology. Moreover, training 3D foundations from scratch is computationally demanding and constrained by limited high-quality volume-text datasets.

In this work, we propose \textbf{Brain3D}, a specialized framework for report generation from volumetric Brain MRI. Rather than relying on slice decomposition or computationally intensive generalist 3D backbones, we adapt a 2D medical encoder via inflation to extract native spatial features. We further identify a crucial bottleneck in the tendency of VLMs to generate verbose, ``caption-like'' descriptions rather than factual diagnostic reports. We address this via a \emph{Staged Vision-Language Grounding} protocol. By progressively moving from latent contrastive alignment to supervised projector \emph{warmup}, and finally to Low-Rank Adaptation (LoRA) \cite{hu2022lora} of the LLM, we guide the model from generic visual recognition to expert neuroradiological syntax. Our contribution is threefold. First, we introduce an \emph{Inflated Volumetric Architecture}, an efficient 3D adaptation of 2D Vision Encoder that enables native spatial processing. Second, we validate a three-stage learning strategy, demonstrating its necessity for minimizing hallucinations and achieving optimal specificity on healthy controls, thereby overcoming a historical limitation of generative VLMs. Finally, we propose a new benchmark for \emph{Clinical Efficacy}, achieving a +130\% gain in the \emph{Clinical Pathology F1 score} over 2D and generalist baselines (0.951 vs.\ 0.413), confirming that volumetric modeling is a necessary condition for diagnostic factualness.

\section{Related Work}
\label{sec:related_work}

We review prior work along three axes: (i) medical report generation with 2D VLMs, (ii) volumetric 3D multimodal models and neuroradiology-specific challenges, and (iii) efficient transfer from 2D foundations to 3D via inflation.\newline

\noindent \textbf{Automated Medical Report Generation.}
Early report generation systems relied on encoder–decoder pipelines, where Convolutional Neural Networks (CNNs) extracted visual features and Long Short-Term Memory (LSTM) networks generated text \cite{yuan2019automatic, wang2021self}, often producing repetitive and locally coherent outputs. 
The advent of the Transformer architecture revolutionized the field by enabling global dependency modeling. Recent Medical Vision–Language Models (VLMs) leverage pretrained Large Language Models (LLMs) to enhance coherence and clinical accuracy. For instance, R2GenGPT \cite{wang2023r2gengpt} introduces a visual alignment module, Med-Flamingo \cite{moor2023med} extends OpenFlamingo for medical few-shot learning, and LLaVA-Med \cite{liu2023visual} adapts instruction tuning to the clinical domain.
\emph{Despite strong linguistic fluency, most state-of-the-art VLMs operate natively on 2D images. When applied to volumetric MRI, they require slice-wise decomposition, potentially disrupting 3D spatial coherence and impairing neuroradiological reasoning.} \newline 

\noindent \textbf{Volumetric 3D Multimodal Models and Neuroradiology.}
Native 3D VLMs have emerged to overcome slice-based limitations. CT2Rep \cite{hamamci2024ct2rep} and CT-CHAT \cite{hamamci2024developing} extend multimodal frameworks to 3D chest CT, while M3D-LaMed \cite{m3d_lamed} proposes a generalist 3D multimodal LLM with token compression strategies. However, these systems are typically general-purpose or CT-focused.
Brain MRIs pose distinct challenges: hyperintense lesion patterns, periventricular signal alterations, hemispheric symmetry, and infiltration topology require coherent volumetric reasoning \cite{Filippi2016MAGNIMS}. Generic pooling or slice aggregation may fragment lesion topology and degrade laterality consistency \cite{Bink2018StructuredReporting}. \emph{Thus, current 3D VLMs work poorly on brain MRIs}. \newline

\noindent \textbf{Transferring 2D Foundations to 3D via Inflation.}
Training large 3D encoders from scratch is computationally demanding. Inflation strategies (I3D) \cite{carreira2017quo}, which extend 2D kernels along the depth axis, provide an efficient alternative and have shown effectiveness in video and medical imaging \cite{merlin}. This paradigm preserves pretrained inductive biases from 2D models (e.g., SigLIP \cite{siglip}, MedGemma \cite{sellergren2025medgemma}) while enabling volumetric processing. \emph{However, current state-of-the-art approaches primarily emphasize architectural transfer and scalability, without explicitly ensuring tight alignment between volumetric spatial grounding and clinically structured language generation. This limitation may hinder consistent reasoning and fine-grained anatomical localization in neuroradiological report generation.} \newline

\noindent \textbf{Our Contribution.} Motivated by these gaps, we introduce \textbf{Brain3D}, a domain-specific framework that combines 3D weight inflation with staged vision–language alignment. This design explicitly separates volumetric grounding from linguistic adaptation for structured brain MRI report generation.

\begin{figure*}[t]
\centering
\includegraphics[width=0.99\textwidth]{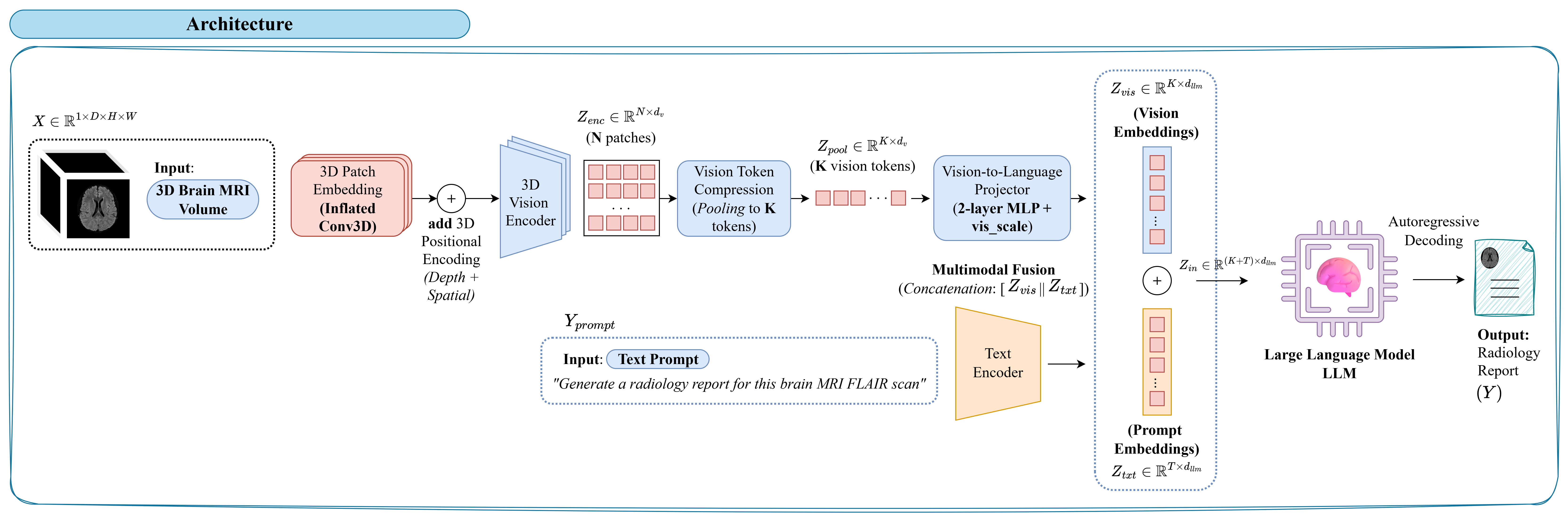}
\caption{\textbf{Brain3D Architecture.}
A standardized MRI volume $X$ is processed by an inflated 3D Transformer
encoder, producing $N$ volumetric patch tokens $Z_{\text{enc}}$.
These tokens are compressed via \emph{Vision Token Compression} into a fixed
set of $K=32$ tokens $Z_{\text{cmp}}$.
The compressed visual tokens are projected into the language embedding space
($Z_{\text{proj}}$) and scaled to obtain conditioning tokens $Z_{\text{cond}}$,
which are prepended to the textual embeddings and used to guide autoregressive
report generation by the causal LLM.}
\label{fig:architecture}
\end{figure*}

\section{Methodology}
\label{sec:methodology}

We introduce \textbf{Brain3D}, a multimodal architecture specifically designed for automated radiology report generation from volumetric brain MRI scans.
The framework adapts a pretrained 2D vision encoder to a native 3D encoder via a weight \emph{inflation} strategy \cite{carreira2017quo} and aligns it to a causal language model through a progressive three-stage vision–language alignment pipeline.

\subsection{Task Formulation}
Let $X \in \mathbb{R}^{C \times D \times H \times W}$ denote a volumetric brain MRI scan, where $C$ represents the intensity channel (for grayscale MRI $C=1$), and $D$, $H$, and $W$ correspond to the depth, height, and width of the resampled volume, respectively.
Let $Y = (y_1, \ldots, y_T)$ be the associated radiology report and $Y_{prompt} = (t_1, \dots, t_S)$ the instruction prompt, represented as a sequence of $T$ and $S$ discrete tokens, respectively, drawn from the language model vocabulary $\mathcal{V}$.
Report generation is formulated as conditional autoregressive decoding:
\begin{equation}
p(Y \mid X, Y_{prompt}) = \prod_{t=1}^{T} p(y_t \mid X, t_1, \dots, t_S, y_{<t}). \label{eq:autoregressive}
\end{equation}
where $y_{<t}$ denotes the previously generated token prefix.

\subsection{Volumetric Data Preprocessing}
To standardize heterogeneous MRI inputs ($X_{\text{raw}}$), we apply skull-stripping, canonical reorientation (RAS) \cite{viswan2023optimizing}, percentile-based intensity clipping ($1^{\text{st}}$-$99^{\text{th}}$) to the $[0, 1]$ interval, and resampling to a fixed grid of $64 \times 128 \times 128$, yielding a final MRI $X_{prep} \in \mathbb{R}^{1 \times 64 \times 128 \times 128}$.

\subsection{Model Architecture}
The model architecture is composed of three functional modules: (i) an inflated 3D vision encoder, (ii) a token compression and projection mechanism, and (iii) a Large Language Model (LLM). The overall pipeline is illustrated in Fig. \ref{fig:architecture}.

\subsubsection{3D Vision Encoder via Inflation}
We initialize the vision backbone from a 2D Transformer pretrained on medical image–text pairs and extend it to volumetric inputs using weight \emph{inflation} \cite{carreira2017quo}. 
Given a 2D patch embedding kernel $W_{2D}$ trained on RGB inputs, we adapt it to single-channel MRI volumes by collapsing the input channel dimension and replicating the kernel along the depth axis to obtain a 3D kernel $W_{3D}$. Inflated weights are normalized to preserve activation scale, enabling volumetric feature extraction while retaining pretrained inductive biases.
To incorporate spatial awareness in 3D, we replace 2D positional embeddings with a decomposed formulation:
\begin{equation}
P_{3D}(z,y,x) = P_{\text{depth}}(z) + P_{\text{spatial}}(y,x),
\end{equation}
where $P_{\text{depth}}$ is learnable and $P_{\text{spatial}}$ reuses pretrained 2D embeddings broadcast along depth.
The final output of the encoder is a sequence of volumetric patch embeddings $Z_{enc} \in \mathbb{R}^{N \times d_v}$, where $N$ is the total number of patches, and $d_v$ is the hidden dimension of the Vision Transformer.

\subsubsection{Visual Token Compression}
Processing the full sequence $Z_{enc}$ (all $N$ volumetric tokens) would be computationally expensive for the LLM. We therefore apply adaptive average pooling along the sequence dimension, producing a fixed set of $K$ visual tokens (with $K=32$ in our setup):
\begin{equation}
Z_{pool} = \mathrm{AdaptiveAvgPool1D}(Z_{enc}) \in \mathbb{R}^{K \times d_v}.
\end{equation}
This operation decouples the volumetric resolution from the LLM context length.

\subsubsection{Vision-Language Projection}
To align visual features ($d_v$) with the LLM embedding space ($d_{llm}$), we use a two-layer MLP with GELU activation. A learnable scalar gate $s$ (\texttt{vis\_scale}) modulates the strength of visual conditioning:
\begin{equation}
Z_{vis} = s \cdot \mathrm{MLP}(Z_{pool}) \in \mathbb{R}^{K \times d_{llm}}.
\end{equation}
A lower scalar $s$ enables gradual visual conditioning of the LLM during training.

\subsubsection{Textual Embedding Extraction}
In parallel, the instruction prompt $Y_{prompt}$ is tokenized as $t$ and then mapped into embeddings via the LLM input matrix as: $Z_{txt} = \text{Embedding}(t) \in \mathbb{R}^{T \times d_{llm}}$.

\subsubsection{LLM Conditioning and Generation}
Instead of using additional cross-attention layers, we adopt a \emph{soft-prompting} approach. The projected visual tokens $Z_{vis}$ are prepended directly to the text embeddings $Z_{txt}$ in the input sequence: $Z_{in} = \mathrm{Concat}(Z_{vis}, Z_{txt}) \in \mathbb{R}^{(K + T) \times d_{llm}}$.
The LLM then performs autoregressive decoding to generate the report $Y$ conditioned on $Z_{\text{in}}$.

\begin{figure*}[t]
    \centering
    \includegraphics[width=\textwidth]{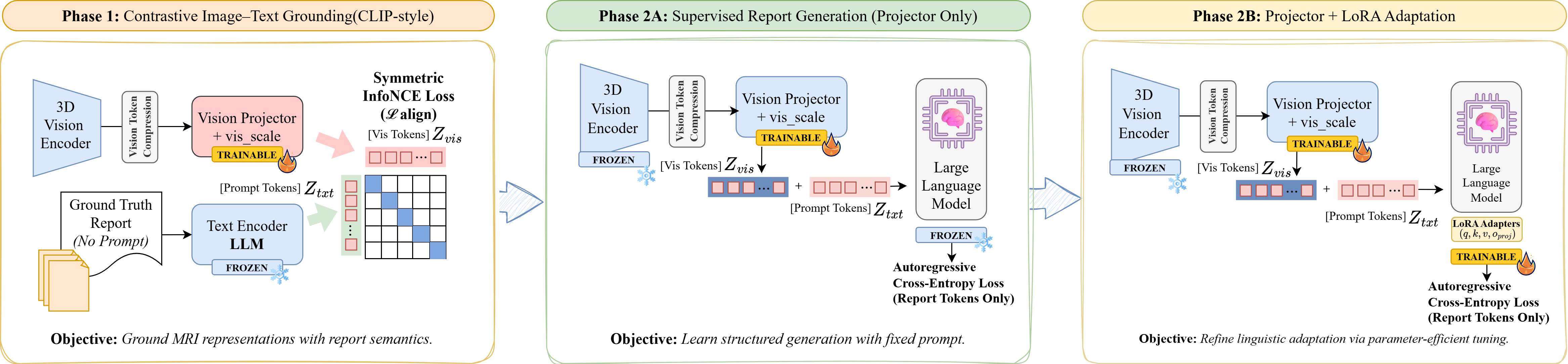}
    \caption{\textbf{Staged Training Strategy} The framework employs a progressive three-phase alignment pipeline. \textbf{Phase 1:} Contrastive Image-Text Grounding aligns the 3D representations ($Z_{vis}$) with report semantics ($Z_t$) using a symmetric InfoNCE loss. \textbf{Phase 2A:} Projector Warmup performs supervised generation with a frozen LLM to stabilize the visual-language mapping. \textbf{Phase 2B:} Linguistic Adaptation fine-tunes the projector and LoRA adapters jointly to capture neuroradiology syntax.
    \textit{Legend:} Modules marked with \textbf{Ice} (\faSnowflake) are frozen; modules marked with \textbf{Fire} (\faFire) are trainable.}
    \label{fig:training_pipeline}
\end{figure*}

\subsection{Staged Vision-Language Alignment}
Training is performed in three stages: (i) contrastive grounding, (ii) supervised projector training, and (iii) supervised fine-tuning with LoRA. (see Fig. \ref{fig:training_pipeline})

\noindent \textbf{Prompting Policy.} We use a single immutable canonical instruction prompt: ``\texttt{Generate a radiology report for this brain MRI FLAIR scan}'' during all training process. During Phase 1 alignment, no prompt is prepended to the report (as shown in Fig. \ref{fig:training_pipeline}).

\subsubsection{Phase 1: Image-Text Grounding via Contrastive Learning}
Phase 1 aligns visual and textual representations via a symmetric bidirectional InfoNCE \cite{chen2020simple}, averaging the image-to-text ($\mathcal{L}_{v \to t}$) and text-to-image ($\mathcal{L}_{t \to v}$) losses. The LLM and vision backbone are frozen; gradients update only the inflated 3D patch embedding $P_{3D}$, MLP projector $\theta_{\text{proj}}$, and scalar $s$.
Let $\mathbf{v}_i$ and $\mathbf{t}_i$ denote $L_2$-normalized global visual and textual embeddings for sample $i$ in a batch of size $B$. We minimize:
\begin{equation}
\mathcal{L}_{Phase_1} = \frac{1}{2} \left( \mathcal{L}_{v \to t} + \mathcal{L}_{t \to v} \right), \text{ where } \mathcal{L}_{v \to t} = \mathcal{L}_{\text{InfoNCE}}(v, t).
\end{equation}
This stage establishes a shared multimodal embedding space prior to generative training.

\subsubsection{Phase 2A: Projector Warmup (Supervised Generation)}
In this phase, with the vision encoder and LLM frozen, we optimize only the MLP projector $\theta_{\text{proj}}$ and the gate $s$ using masked next-token prediction \cite{gloeckle2024better}.
We construct the input sequence $U$ by concatenating the visual tokens $Z_{vis}$, the canonical prompt tokens $Z_{txt}$, and the ground-truth report tokens $Z_Y$ as: $U = \mathrm{Concat}(Z_{vis}, Z_{txt}, Z_Y)$.
The loss is computed using a binary mask $M$ to ensure gradients are calculated only on the report tokens, effectively ignoring the visual and prompt tokens (set to $-100$ in our implementation). We minimize:
\begin{equation}
\mathcal{L}_{Phase_{2A}} = M_t \cdot \mathcal{L}_{next\_token}(U; \theta_{proj}, s).
\end{equation}
This ``warmup'' stage stabilizes visual conditioning before adapting the language model.

\subsubsection{Phase 2B: Linguistic Fine-Tuning with LoRA}
After projector stabilization, we freeze the 3D Vision Encoder and jointly optimize the MLP projector $\theta_{\text{proj}}$ and LoRA adapters \cite{hu2022lora} injected into LLM attention layers.
The optimization objective remains the masked next token prediction loss of Phase 2A, now jointly optimized over the Projector $\theta_{proj}$ and LoRA parameters $LoRA_{param}$, as follows:
\begin{equation}
\mathcal{L}_{Phase_{2B}} = M_t \cdot \mathcal{L}_{next\_token}(U; \theta_{proj}, LoRA_{param}).
\end{equation}
This final stage adapts the linguistic space to neuroradiological reporting while preserving volumetric grounding.

\subsection{Inference Strategy}
During inference, the model operates in autoregressive generation mode as described in Eq. \ref{eq:autoregressive}.
We employ conservative stochastic decoding (temperature $T=0.1$, top-$p=0.9$) with repetition penalty ($\theta=1.2$) and trigram blocking to prevent redundancy while maintaining diagnostic stability.

\section{Experiments}
\label{sec:experiments}

\subsection{Experimental Setup}
Experiments were conducted on a single NVIDIA A100 (64GB VRAM) using PyTorch and HuggingFace frameworks.
We initialize the 2D vision backbone from \texttt{MedSigLIP} \cite{sellergren2025medgemma} and adopt \texttt{MedGemma 1.5-4B-IT}\footnotemark[2] as the causal language model (LLM) for text generation.
Training employed bfloat16 mixed precision with gradient accumulation (effective batch size 128). Optimization used AdamW with linear warmup and cosine decay; early stopping was applied after 15 epochs without validation improvement. LoRA \cite{hu2022lora} was configured with rank $r=16$ and scaling factor $\alpha=32$.

\subsection{Datasets and Preprocessing}
We constructed a 3D dataset comprising pathological cases and healthy controls to model both tumoral and normal brain anatomy. The final dataset includes $N=468$ subjects, detailed as follows:
\textbf{Pathological Cohort (BraTS).} We selected 369 FLAIR volumes from the \textit{BraTS2020} training set \cite{Menze2015BRATS}, using TextBraTS annotations \cite{shi2025textbratstextguidedvolumetricbrain} to derive structured reports (location, edema, necrosis). Laterality distribution is balanced (42.5\% left, 40.7\% right, 14.6\% bilateral; 2.2\% undefined).
\textbf{Healthy Controls.} We included 99 healthy brain MRIs from OpenNeuro/Brainlife \cite{brainlife_sani} (21.2\% of total dataset) to prevent pathological bias and allow the model to internalize the representation of healthy anatomy. We performed strict subject-level splitting (70/10/20 train/val/test) to prevent data leakage, stratified by class and lesion laterality.


\subsection{Baselines}
To validate our model, we compare against state-of-the-art 2D and 3D medical VLMs to isolate the impact of native volumetric encoding.
We evaluate: \texttt{MedGemma 1.5-4B-IT}\footnote{\url{https://huggingface.co/google/medgemma-1.5-4b-it}} \cite{sellergren2025medgemma}, a 2D medical VLM adapted via a sequence-based strategy, where all 64 axial slices are processed as independent images within the context window. This baseline evaluates whether a powerful 2D model can implicitly learn 3D spatial relationships solely through prompt sequencing, or if explicit volumetric encoding (as in our approach) is required for accurate lesion localization;
\texttt{Med3DVLM}\footnote{\url{https://huggingface.co/MagicXin/Med3DVLM-Qwen-2.5-7B}} \cite{med3dvlm}, a generalist 3D VLM pretrained on large-scale M3D-Data. This comparison isolates the contribution of domain-specific staged alignment beyond generic 3D pretraining.

\subsection{Evaluation Metrics}
We report all metrics with 95\% confidence intervals. Linguistic quality and similarity is assessed against ground truth references using BLEU-1/4 \cite{papineni2002bleu}, ROUGE-1/2/L \cite{lin2004rouge}, METEOR \cite{banerjee2005meteor}, BERTScore \cite{zhang2019bertscore}, and CIDEr \cite{vedantam2015cider}. 
However, these metrics often fail to capture factual medical correctness (e.g., misidentifying the lesion side minimally affects BLEU but constitutes a critical clinical error). Therefore, we implemented a rule-based extraction module to calculate the F1-score for the following specific clinical categories: \textbf{Clinical Laterality F1:} Measures accuracy in detecting the lesion side (Left, Right, Bilateral); \textbf{Clinical Anatomy F1:} Evaluates specific anatomical localization (e.g., Frontal, Parietal, Ventricle); \textbf{Clinical Pathology F1:} Assesses the correct identification of pathological descriptors (e.g., Edema, Necrosis, Enhancement, Compression).

\begin{table*}[!t]
\centering
\caption{\textbf{Main Results \& Ablation.} 
Top: Comparison against Med3DVLM and MedGemma 1.5 4B baselines. Bottom: Evolution of our framework through training phases.
\textbf{Note the massive gain in Clinical Efficacy (F1)} of our final model (Phase 2b) compared to the strong 2D baseline (MedGemma 1.5).
Best results per column are in \textbf{bold}. Values: Mean \scriptsize{(95\% CI)}.}
\label{tab:main_results}

\resizebox{\textwidth}{!}{%
\begin{tabular}{l cccccc cc ccc}
\toprule

\multirow{2}{*}{\textbf{Model}} & 
\multicolumn{6}{c}{\textbf{Natural Language Generation (NLG)}} & 
\multicolumn{2}{c}{\textbf{Semantics}} & 
\multicolumn{3}{c}{\textbf{Clinical Efficacy (F1)}} \\
\cmidrule(lr){2-7} \cmidrule(lr){8-9} \cmidrule(lr){10-12}

 & \textbf{B-1} & \textbf{B-4} & \textbf{R-1} & \textbf{R-2} & \textbf{R-L} & \textbf{MTR} 
 & \textbf{BERT} & \textbf{CIDEr} 
 & \textbf{Lat} & \textbf{Anat} & \textbf{Path} \\
\midrule

\multicolumn{12}{l}{\textit{\textbf{State-of-the-Art Baselines}}} \\
\addlinespace[0.1cm]

\textbf{Med3DVLM} \cite{med3dvlm} & 
0.051 & 0.005 & 0.108 & 0.018 & 0.083 & 0.055 & 
0.836 & 0.007 & 
0.300 & 0.225 & 0.119 \\
\textit{(3D Generalist)} & 
\scriptsize{[0.04, 0.06]} & \scriptsize{[0.00, 0.01]} & \scriptsize{[0.09, 0.13]} & \scriptsize{[0.01, 0.02]} & \scriptsize{[0.07, 0.10]} & \scriptsize{[0.05, 0.06]} & 
\scriptsize{[0.83, 0.84]} & \scriptsize{[0.00, 0.00]} & 
\scriptsize{[0.22, 0.40]} & \scriptsize{[0.15, 0.30]} & \scriptsize{[0.08, 0.18]} \\
\addlinespace[0.1cm]

\textbf{MedGemma 1.5} \cite{sellergren2025medgemma} & 
0.245 & 0.024 & 0.326 & 0.077 & 0.189 & 0.190 & 
0.859 & 0.029 & 
0.526 & 0.461 & 0.413 \\
\textit{(2D Slice-based)} & 
\scriptsize{[0.22, 0.27]} & \scriptsize{[0.02, 0.03]} & \scriptsize{[0.30, 0.35]} & \scriptsize{[0.07, 0.09]} & \scriptsize{[0.17, 0.20]} & \scriptsize{[0.18, 0.20]} & 
\scriptsize{[0.86, 0.86]} & \scriptsize{[0.02, 0.03]} & 
\scriptsize{[0.45, 0.60]} & \scriptsize{[0.39, 0.53]} & \scriptsize{[0.36, 0.46]} \\

\midrule

\multicolumn{12}{l}{\textit{\textbf{Brain3D Framework (Ours)}}} \\
\addlinespace[0.1cm]

\textit{Phase 1: Alignment} & 
0.122 & 0.005 & 0.177 & 0.018 & 0.113 & 0.128 & 
0.800 & 0.003 & 
0.243 & 0.141 & 0.211 \\
\textit{(Zero-shot / No Projector)} & 
\scriptsize{[0.11, 0.13]} & \scriptsize{[0.00, 0.01]} & \scriptsize{[0.17, 0.19]} & \scriptsize{[0.02, 0.02]} & \scriptsize{[0.11, 0.12]} & \scriptsize{[0.12, 0.14]} & 
\scriptsize{[0.79, 0.81]} & \scriptsize{[0.00, 0.00]} & 
\scriptsize{[0.16, 0.32]} & \scriptsize{[0.08, 0.21]} & \scriptsize{[0.15, 0.28]} \\
\addlinespace[0.1cm]

\textit{Phase 2a: Warmup} & 
0.280 & \textbf{0.099} & 0.382 & \textbf{0.158} & 0.285 & 0.250 & 
0.884 & \textbf{0.504} & 
0.658 & 0.503 & 0.711 \\
\textit{(Projector Trained Only)} & 
\scriptsize{[0.25, 0.32]} & \scriptsize{[0.07, 0.14]} & \scriptsize{[0.35, 0.42]} & \scriptsize{[0.13, 0.20]} & \scriptsize{[0.25, 0.32]} & \scriptsize{[0.21, 0.29]} & 
\scriptsize{[0.88, 0.89]} & \scriptsize{[0.49, 0.51]} & 
\scriptsize{[0.58, 0.74]} & \scriptsize{[0.43, 0.57]} & \scriptsize{[0.66, 0.76]} \\
\addlinespace[0.1cm]

\rowcolor{gray!10}
\textbf{Phase 2b: Full Model} & 
\textbf{0.302} & 0.098 & \textbf{0.407} & 0.155 & \textbf{0.289} & \textbf{0.253} & 
\textbf{0.898} & 0.293 & 
\textbf{0.689} & \textbf{0.691} & \textbf{0.951} \\
\rowcolor{gray!10}
\textit{(Projector + LoRA)} & 
\scriptsize{[0.28, 0.33]} & \scriptsize{[0.08, 0.13]} & \scriptsize{[0.38, 0.44]} & \scriptsize{[0.13, 0.19]} & \scriptsize{[0.27, 0.32]} & \scriptsize{[0.23, 0.28]} & 
\scriptsize{[0.89, 0.90]} & \scriptsize{[0.29, 0.30]} & 
\scriptsize{[0.61, 0.76]} & \scriptsize{[0.64, 0.74]} & \scriptsize{[0.91, 0.98]} \\

\bottomrule
\end{tabular}
}
\end{table*}

\subsection{Quantitative Analysis and Ablation Study}
Tab. \ref{tab:main_results} highlights a clear divergence between linguistic fluency and clinical correctness. \texttt{MedGemma 1.5} achieves strong semantic similarity (BERTScore 0.859) but low Clinical Pathology F1 (0.413), reflecting slice-based spatial inconsistencies and laterality errors. \texttt{Med3DVLM} underperforms on both linguistic and clinical metrics, indicating that generic 3D pretraining alone is insufficient for neuroradiological reporting.
In contrast, our final model (Phase 2b) achieves 0.951 Clinical Pathology F1 (+130\% over \texttt{MedGemma 1.5}), confirming that native volumetric encoding combined with staged alignment is necessary for robust spatial grounding.
The ablation study reveals progressive specialization: \textit{Phase 1 (Latent Alignment)} establishes latent alignment but is not optimized for generation (low NLG scores); \textit{Phase 2a (Projector Warmup)} maximizes descriptive fluency (CIDEr: 0.504; ROUGE-L: 0.285); Phase 2b shifts toward structured, clinically precise reporting, sacrificing caption-style verbosity for factual accuracy (Clinical Pathology F1: 0.951).

\setlength{\tabcolsep}{3pt} 

\begin{table*}[t]
\centering
\caption{\textbf{Qualitative Comparison.} Visual comparison of generated reports for a representative test sample. Our model identifies the lesion location and pathologies, whereas baselines hallucinate or fail.}
\label{tab:qualitative_comparison}

\renewcommand{\arraystretch}{1.25}

\begin{tabularx}{\textwidth}{
|m{1mm}
>{\centering\arraybackslash}m{0.13\textwidth}
m{1mm}
|X|}
\hline
\rowcolor{headergray}
\multicolumn{4}{|c|}{\textbf{Qualitative Comparison of Generated Reports}} \\
\hline

&
\multirow{4}{*}{
  \adjustbox{valign=m}{
    \includegraphics[width=0.90\linewidth]{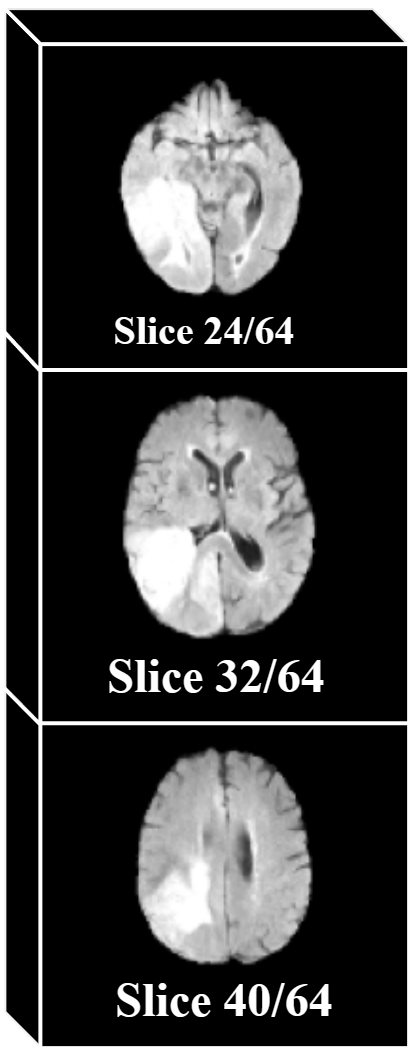}
  }
}
&
&
\textbf{$<$Ground Truth$>$}: The lesion area is in the \textbf{frontal and parietal regions of the left hemisphere} with mixed signal intensities of varying levels and speckled high-intensity signal areas. \textbf{Edema} is primarily concentrated in the peripheral regions of the left frontal and parietal lobes, indicating moderate to severe swelling of the tissues surrounding the lesions. \textbf{Necrosis} is observed in some lesion areas displaying low signal intensity interspersed within regions of high and isointense signals, showing a certain mottled appearance. \textbf{Ventricular compression} is mild deformation of the ventricles, suggesting pressure effects of the lesions on these regions. \\
\cline{4-4}

&
&
&
\textbf{$<$Brain3D (Ours)$>$}: The lesion area is in the \textcolor{correctgreen}{\textbf{left parietal and frontal lobes}} with mixed high-signal areas. \textcolor{correctgreen}{\textbf{Edema signals are mainly observed}} around these lesions, indicating significant edema presence affecting parts of both frontal and temporal regions as well as some portions within the parietal lobe. \textcolor{correctgreen}{\textbf{Necrosis may be present}} at low signal intensity or scattered throughout certain sections of the brain tissue affected by edema but shows no clear concentration. \textcolor{correctgreen}{\textbf{Ventricular compression effects}} on adjacent ventricles can occur due to pressure from surrounding tissues near the ventricular system. \\
\cline{4-4}

&
&
&
\textbf{$<$MedGemma Baseline$>$}: There is a large, irregular, mixed signal intensity mass centered in the \textcolor{wrongred}{\textbf{right frontal lobe}}, extending into the \textcolor{wrongred}{\textbf{right temporal lobe}} and involving the \textcolor{wrongred}{\textbf{right lateral ventricle}}. There is significant \textcolor{wrongred}{\textbf{mass effect on the right lateral ventricle}}, causing mild to moderate effacement of the adjacent sulci. \\
\cline{4-4}

&
&
&
\textbf{$<$Med3D-VLM$>$}: No intracranial hemorrhage. No evidence of acute infarction. \textcolor{wrongred}{\textbf{No mass effect}} or midline shift. \textcolor{wrongred}{\textbf{Normal grey-white matter differentiation}}. \\
\hline

\end{tabularx}
\end{table*}

\begin{figure}[t]
    \includegraphics[width=0.46\textwidth]{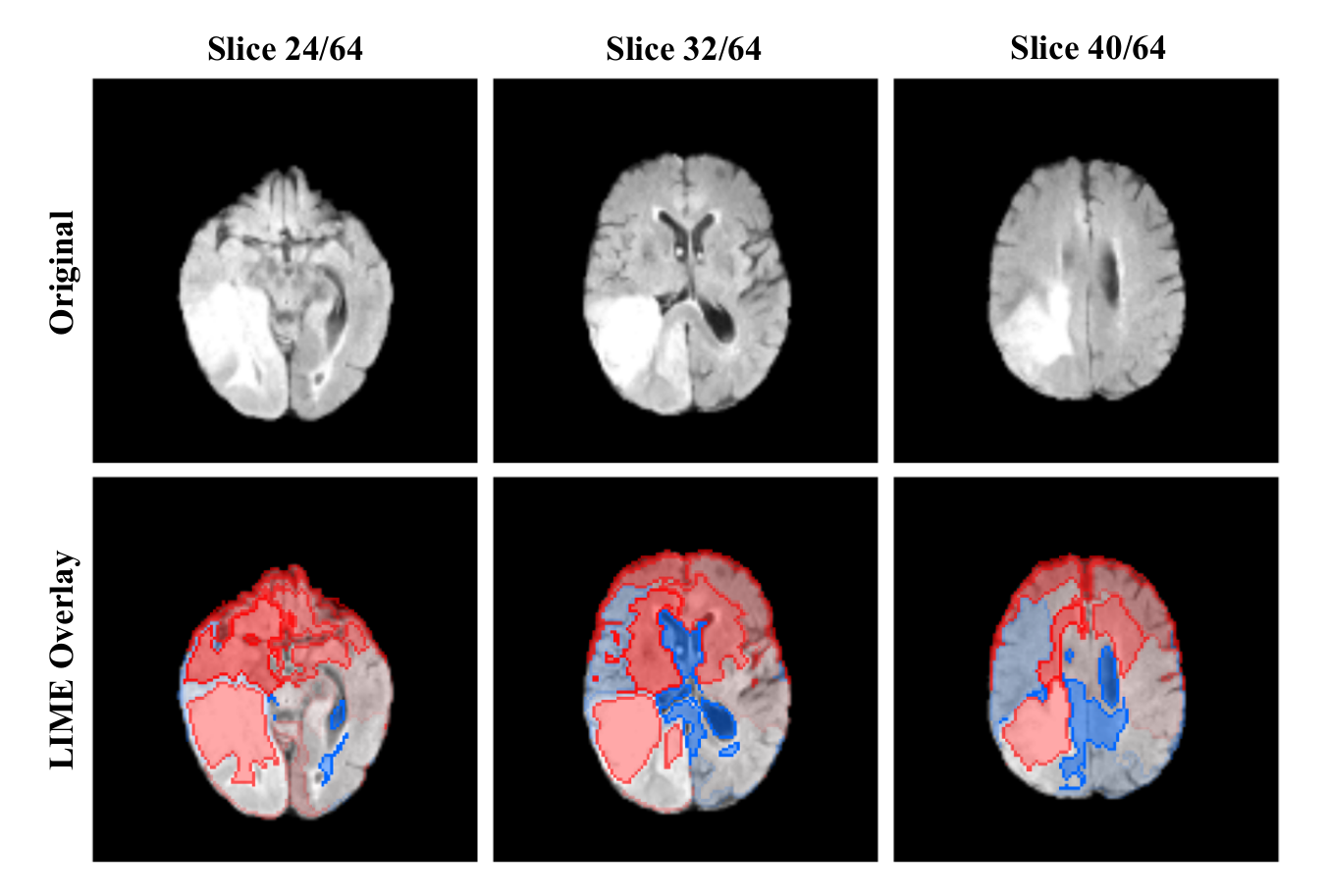}
    \caption{\textbf{3D LIME Attribution Maps.} 
    Volumetric grounding visualized via 3D LIME over SLIC supervoxels for a representative test case. Red regions indicates positive attribution (supporting the report), blue regions negative attribution. The tumor-bearing hemisphere is correctly highlighted; however, diffuse and partially contralateral supervoxels are also activated, suggesting reliance on both lesion-centered and global contextual patterns, potentially contributing to lateralization errors.}
    \label{fig:LIME}
\end{figure}

\subsection{Qualitative Analysis and Interpretability} \label{subsec:qual_analysis}
Beyond quantitative metrics, we conduct a qualitative assessment to evaluate grounding fidelity and systematically characterize residual failure modes.

\noindent \textbf{Qualitative Answer Comparison.} Tab. \ref{tab:qualitative_comparison} shows that slice-based models exhibit spatial inconsistencies, while generalist 3D models lack domain specialization. \textbf{Brain3D} correctly identifies lesion location and pathology, with negligible hallucinations on healthy scans. Residual errors primarily involve edema–necrosis ambiguities in complex cases.

\noindent \textbf{Interpretability (3D LIME over Supervoxels).} We apply 3D LIME over SLIC supervoxels (Fig. \ref{fig:LIME}). This analysis is exploratory and not intended as a definitive attribution study; despite known limitations in high-dimensional settings, LIME provides a lightweight sanity check of lesion-level grounding.
Attribution maps demonstrate predominantly lesion-centered attribution within the tumor-bearing hemisphere. However, occasional diffuse or partially contralateral attribution indicates reliance on both focal and global contextual cues.

\noindent \textbf{Error Analysis.} The most frequent failure mode is laterality inversion ($\approx 15\%$ of pathological cases), where morphology is correct but hemispheric positioning is flipped. In diffuse gliomas, peripheral infiltration may be under-reported. Under uncertainty, mild distributional bias toward frequent anatomical patterns is observed (e.g., “left parietal and occipital lobes”). Overall, errors stem from positional ambiguity rather than random hallucination.

\section{Conclusion and Future Work}
\label{sec:conclusion}

The work presents \textbf{Brain3D}, a volumetric vision-language framework for neuroradiological report generation that bridges the 2D-to-3D adaptation gap through weight inflation and staged alignment.
Unlike slice-based models that achieve high lexical scores yet fail clinically, our approach prioritizes diagnostic factuality, reaching 0.951 Clinical Pathology F1 and perfect specificity on healthy controls. Results demonstrate that decoupling visual grounding from linguistic specialization is critical for reducing hallucinations in medical VLMs.
Future work will investigate anatomically informed positional embeddings to mitigate lateralization errors, correcting distributional bias using DPO/RLHF to encourage accurate spatial descriptions, and scaling pretraining to larger, multi-sequence MRI datasets (T1, T2, FLAIR) to enable broader multi-modal neuroradiology assistance.


\bibliographystyle{IEEEtran}
\bibliography{samplebib.bib}

\end{document}